\documentclass[10pt,twocolumn,letterpaper]{article}

\usepackage[pagenumbers]{cvpr} 

\usepackage{booktabs}

\usepackage[utf8]{inputenc} 
\usepackage[T1]{fontenc}    
\usepackage{url}            
\usepackage{booktabs}       
\usepackage{amsfonts}       
\usepackage{nicefrac}       
\usepackage{microtype}      
\usepackage[table]{xcolor}         
\usepackage{amsmath}
\usepackage{graphicx}
\usepackage{wrapfig}
\usepackage{graphics}
\usepackage{boldline}
\usepackage{multirow}
\usepackage{pifont}
\usepackage{caption}
\usepackage{subcaption}
\usepackage{relsize}
\newcommand{\cx}{\ding{55}}
\newcommand{\cy}{\ding{51}}
\newcommand{\cc}{\ding{61}}

\usepackage[pagebackref,breaklinks,colorlinks]{hyperref}

\usepackage[capitalize]{cleveref}
\crefname{section}{Sec.}{Secs.}
\Crefname{section}{Section}{Sections}
\Crefname{table}{Table}{Tables}
\crefname{table}{Tab.}{Tabs.}

\begin{document}

\title{Soft-Landing Strategy for Alleviating the Task Discrepancy Problem \\ in Temporal Action Localization Tasks}

\author{%
Hyolim Kang$^1$ \quad Hanjung Kim$^1$ \quad Joungbin An$^1$ \quad Minsu Cho$^2$ \quad Seon Joo Kim$^1$   \\
$^1$Yonsei University \quad $^2$POSTECH \\
\texttt{\{hyolimkang, hanjungkim, joungbinan, seonjookim\}@yonsei.ac.kr}\\
\texttt{mscho@postech.ac.kr}
}
\maketitle

\begin{abstract}
Temporal Action Localization (TAL) methods typically operate on top of feature sequences from a frozen snippet encoder that is pretrained with the Trimmed Action Classification (TAC) tasks, resulting in a task discrepancy problem. 
While existing TAL methods mitigate this issue either by retraining the encoder with a pretext task or by end-to-end finetuning,  
they commonly require an overload of high memory and computation.
In this work, we introduce Soft-Landing (SoLa) strategy, an efficient yet effective framework to bridge the transferability gap between the pretrained encoder and the downstream tasks by incorporating a light-weight neural network, i.e., a SoLa module, on top of the frozen encoder.
We also propose an unsupervised training scheme for the SoLa module; it learns with inter-frame Similarity Matching that uses the frame interval as its supervisory signal, eliminating the need for temporal annotations.
Experimental evaluation on various benchmarks for downstream TAL tasks shows that our method effectively alleviates the task discrepancy problem with remarkable computational efficiency.

\end{abstract}
\section{Introduction}
\label{sec:intro}

Our world is full of untrimmed videos, including a plethora of Youtube videos, security camera recordings, and online streaming services.  
Analyzing never-ending video streams is thus one of the most promising directions of computer vision research in this era~\cite{wu2021towards}.
Amongst many long-form video understanding tasks, the task of finding action instances in time and classifying their categories, known as Temporal Action Localization (TAL), has received intense attention from both the academia and the industry in recent years; 
TAL is considered to be the fundamental building block for more sophisticated video understanding tasks since it plays the basic role of distinguishing frame-of-interest from irrelevant background frames~\cite{xu2020gtad, lin2019bmn, zhang2022actionformer}.

\begin{figure}
    \includegraphics[width=\linewidth]{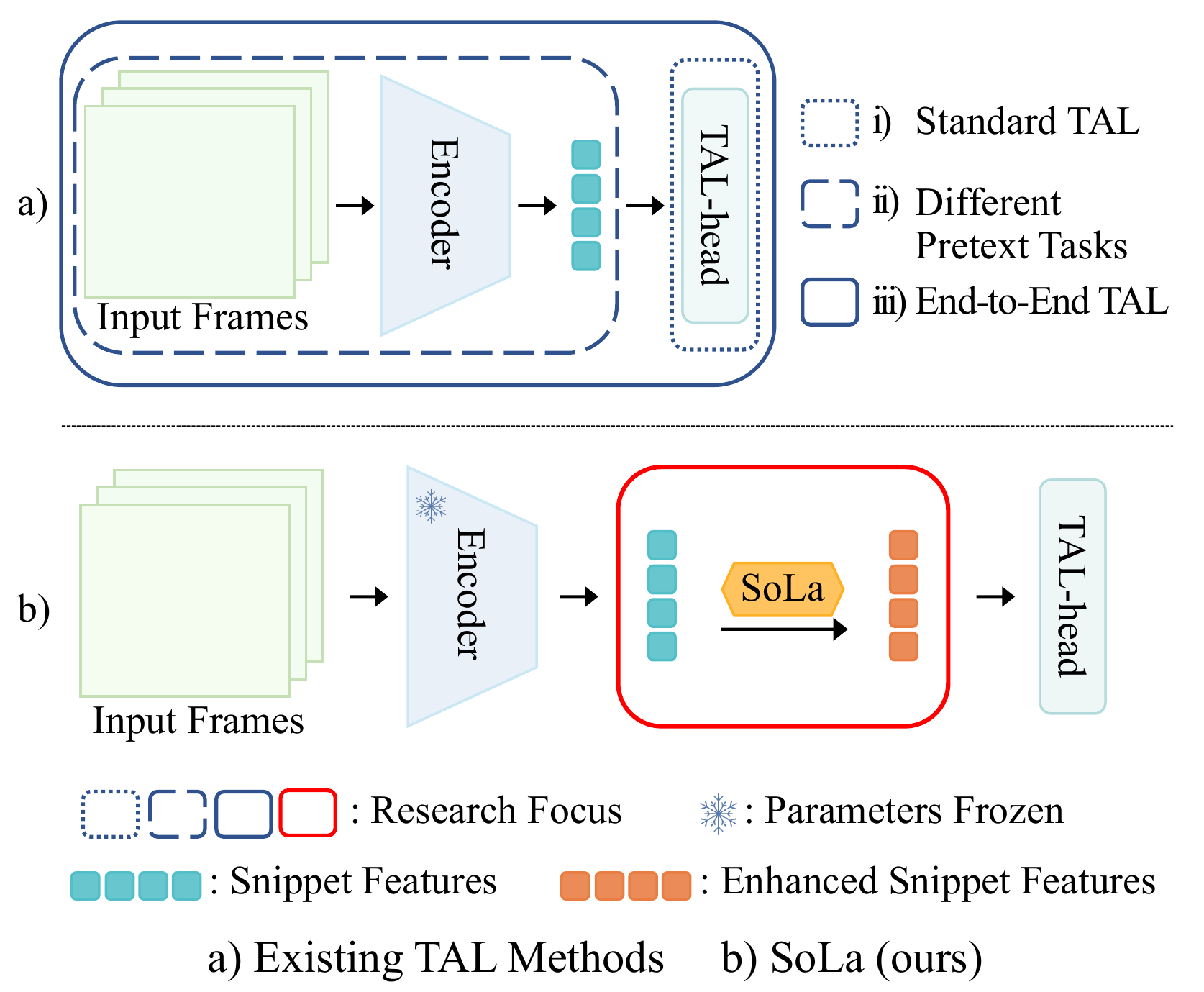}
    \caption{(a-i) Standard TAL framework assumes a ``frozen snippet encoder'' and only focuses on designing a good TAL head, causing the task discrepancy problem. Straightforward approaches to alleviating the issue include devising (a-ii) a temporally sensitive pretext task~\cite{xu2021boundary, zhang2022unsupervised, alwassel2021tsp}, and (a-iii) an end-to-end TAL training procedure~\cite{xu2021low, liu2022empirical}.
    However, both approaches break the aforementioned frozen snippet encoder assumption.
    On the other hand, (b) our SoLa strategy shares the ``frozen snippet encoder assumption'' with the standard TAL framework by providing a \textit{smooth linkage} between the frozen encoder and the downstream head.
    It offers general applicability and more importantly, exceptional computational efficiency.
    }
    \label{fig:teaser}
\end{figure}

Despite its importance, training a TAL model has a unique computational challenge that hinders the naive extension of conventional image processing models, mainly due to the large size of the model input.
For instance, videos in the wild can be several minutes or even hours long, implying that loading the whole video to a device for processing is often infeasible.
In this context, the prevailing convention in processing a long-form video for TAL is to divide the video into non-overlapping short \textit{snippets} and deal with the snippet-wise feature sequences.
Specifically, a standard training pipeline for the long-form video understanding tasks consists of two steps: (i) Train the snippet encoder with a large-scale action recognition dataset (e.g., Kinetics400), which is often different from the dataset for the downstream task;
(ii) Train the downstream head (e.g., TAL) that takes the snippet feature sequences extracted from the pretrained encoder.
An issue here is that the mainstream pretext task for the snippet-wise video encoder is ``Trimmed'' Action Classification (TAC), which does not  handle action boundaries and background frames.
Although the current pipeline achieves remarkable performance in TAL tasks due to the power of large action recognition datasets, recent works ~\cite{xu2021boundary, alwassel2021tsp,xu2021low, zhang2022unsupervised} point out the \textit{task discrepancy} problem that is inherent in this two-staged approach. The task discrepancy problem, first introduced in~\cite{xu2021low}, comes from the pretrained snippet encoder's insensitivity to different snippets within the same action class.
It results in a temporally invariant snippet feature sequence, making it hard to distinguish foreground actions from backgrounds.
A straightforward approach to the problem is adopting a \textit{temporally sensitive pretext task} to train the snippet encoder~\cite{alwassel2021tsp, xu2021boundary}, or devising an end-to-end framework~\cite{xu2021low, liu2022empirical}, which are briefly described in  Figure~\ref{fig:teaser} (a).
However, as all previous methods involve retraining the snippet encoder, an excessive use of memory and computation is inevitable.

To tackle the task discrepancy problem, we propose a new approach, namely \textbf{Soft-Landing} (SoLa) strategy, which is neither memory nor computationally expensive.
SoLa strategy is a novel method which incorporates a light-weight neural network, i.e., Soft-Landing (SoLa) module, between the pretrained encoder and the downstream head.
The properly trained SoLa module will act like a middleware between the pretext and the downstream tasks, mitigating the task discrepancy problem (Figure~\ref{fig:teaser} (b)).
Since the task adaptation is solely done by the SoLa module, the parameters of the pretrained encoder are fixed in our SoLa strategy.
The use of a frozen encoder significantly differentiates our approach from previous methods that mainly focus on designing an appropriate training methodology for a snippet encoder.
In addition, our SoLa strategy only requires an access to the pre-extracted snippet feature sequence, being fully compatible with the prevailing two-stage TAL framework.

We also propose \textbf{Similarity Matching}, an unsupervised training scheme for the SoLa module that involves neither frame-level data manipulation nor temporal annotations. Our training strategy circumvents the need for strong frame-level data augmentation which most existing unsupervised representation learning techniques~\cite{chen2020simple, grill2020bootstrap} rely on. This strategy perfectly suits our condition where frame-level data augmentation is impossible, as we only have an access to pre-extracted snippet features.
The new loss is based on a simple empirical observation: ``adjacent snippet features are similar, while distant snippet features remain distinct''.
Coupled with the Simsiam~\cite{chen2021exploring} framework, Similarity Matching not only prevents the collapse, but also induces temporally sensitive feature sequences, resulting in a better performance in various downstream tasks.

The contributions of the paper can be summarized as follows:
\begin{center}
    \begin{enumerate}
        \item To tackle the task discrepancy problem, we introduce a novel \textbf{Soft-Landing} (SoLa) strategy, which does not involve retraining of the snippet encoder.
        As we can directly deploy the ``frozen'' pretrained snippet encoder without any modification, our SoLa strategy offers easier applicability compared to previous works that require snippet encoders to be retrained. 
        \item We propose \textbf{Similarity Matching}, a new self-supervised learning algorithm for the SoLa strategy.
        As frame interval is utilized as its only learning signal, it requires neither data augmentation nor temporal annotation.
        \item With our SoLa strategy, we show significant improvement in performance for  downstream tasks, outperforming many of the recent works that involve computationally heavy snippet encoder retraining.
    \end{enumerate}
\end{center}
\section{Related Work}
\subsection{Temporal Action Localization Tasks}
As a fundamental task for processing long-form videos, Temporal Action Localization (TAL) has drawn significant attention among computer vision researchers leading to a plethora of seminal works~\cite{xu2020gtad, lin2019bmn, zeng2019graph, chao2018rethinking, lin2020fast, Zhang2021tpami, Xiao2021aaai}.
Beyond the standard fully-supervised and offline setting, various extensions of the task were suggested, including its online~\cite{kang2021cag, kim2022sliding}, weakly-supervised~\cite{ma2020sf, yang2021uncertainty, zhang2021cola} and unsupervised~\cite{Gong2020cvpr} variants.

\subsection{Designing Better Snippet Encoder}
Although numerous attempts have been made for designing better TAL heads, relatively less attention has been paid for devising a \textit{good snippet encoder}, despite the fact that all TAL methods and its variants start from the pre-extracted snippet feature sequences.
A major problem of the TAC-pretrained snippet encoder is its insensitivity to different snippets in the same video clip.
Apparently, this insensitivity problem can be resolved by adopting a ``temporally sensitive'' pretraining method for the snippet encoder.
In this perspective,~\cite{alwassel2021tsp} rigorously exploited temporal annotations for training the snippet encoder.
However, as it requires a large scale and \textit{temporally annotated} video dataset,  general applicability of this approach is limited.
On the other hand, \cite{xu2021boundary} adopted data-generation approach which only exploited the action class labels of the Kinetics400~\cite{kay2017kinetics} dataset.
To be specific, various boundaries were generated by simply stitching intra-/inter- class videos, and a pretext task of guessing the type of the generated boundaries was proposed.
Going one step further, a recent work \cite{zhang2022unsupervised} introduced a completely \textit{unsupervised} pretraining methodology for the snippet encoder, greatly expanding its possible applications.
In addition, it is worth noting that \cite{xu2021low} made an initial attempt on designing an end-to-end TAL with a low-fidelity snippet encoder, while \cite{liu2022empirical} provided exhaustive empirical studies on these end-to-end TAL approaches.
Nevertheless, all previous works involve \textit{trainable snippet encoder} assumption, while our SoLa strategy only requires pre-extracted feature sequence for its adoption.

\subsection{Temporal Self-similarity Matrix}
Temporal Self-similarity Matrix (TSM) is a square matrix, where each of its element corresponds to its self-similarity score.
From a given video with $L$ frames, each value at position $(i, j)$ in TSM is calculated using a similarity metric (e.g. cosine similarity) between frame i and j, resulting in an $L \times L$ matrix.
As an effective and interpretable intermediate representation of a given video, several recent works~\cite{kang2021uboco, kang2021winning} exploited TSM to tackle various video understanding tasks, including Generic Event Boundary Detection~\cite{shou2021generic} and  repetitive action counting~\cite{dwibedi2020counting}. 
In our work, we focus on the certain similarity patterns that arise in general TSM, motivating us to design a new Similarity Matching objective.

\section{Method}

\subsection{Problem Description}
Let $V:=v_{\psi=1}^{l}$ be a video consisting of $l$ frames.
We assume having a pretrained snippet encoder that takes a length $\alpha$ snippet $v_{\alpha\psi +1 }^{\alpha \psi +\alpha}$ as its input and emits a vector $f\in \mathbb{R}^{m}$.
With the snippet encoder, we convert $V$ into a snippet feature sequence $f_{\tau =1}^{L}$, where $L=\lceil l/\alpha \rceil$ if there is no overlap among the snippets.
Here, we introduce the Soft-Landing (SoLa) module $SoLa(\cdot) : \mathbb{R}^{L \times m} \to \mathbb{R}^{L \times m}$, where $F_{\tau =1}^{L} = SoLa(f_{\tau =1}^{L})$ and $F_{\tau =1}^{L}$ denotes the transformed feature sequence\footnote{In general, output dimension of the $SoLa(\cdot)$ can be different. But here we only consider the same dimension case for clarity.}.
For a wider and more general applicability of the transformed feature sequence $F_{\tau =1}^{L}$, we assume to have access to the \textit{unlabeled} downstream dataset, meaning that we only know the target data but do not know the target task.
Our ultimate goal is to devise a proper unsupervised training method to train the SoLa module that produces temporally sensitive transformed feature sequence $F_{\tau =1}^{L}$.
We expect the transformed feature sequence $F_{\tau =1}^{L}$ to perform better than the original feature sequence $f_{\tau =1}^{L}$ in various downstream tasks, in which the temporal sensitivity is important.

\begin{figure}[t]
        \includegraphics[width=\linewidth]{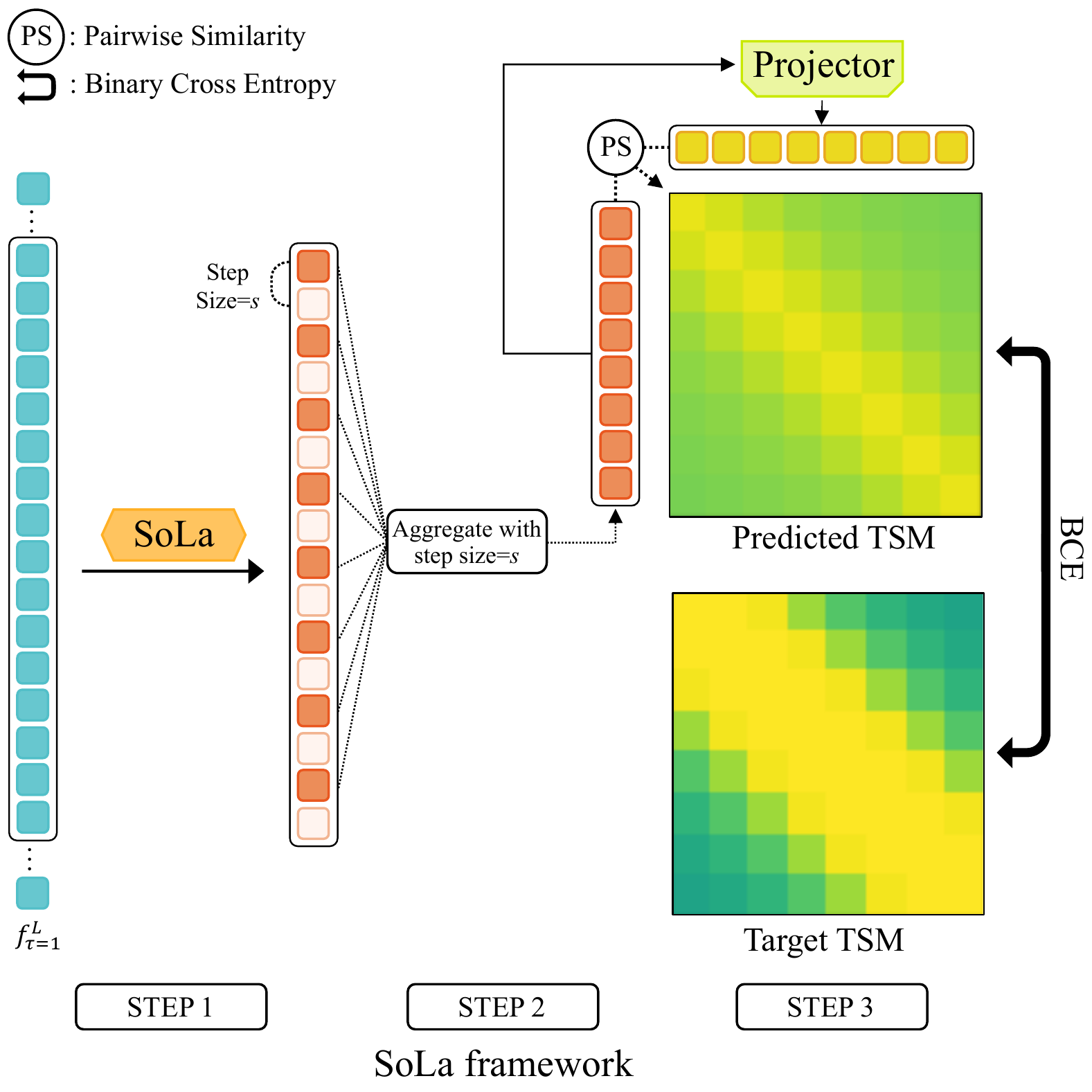}
    \caption{
        Step by step instruction of the overall SoLa module training:
        [Step 1] A fixed size subsequence from the snippet feature sequence $f_{\tau =1}^{L}$ is sampled, which is then passed through the SoLa module. 
        [Step 2] Each transformed feature is aggregated with a predefined step size $s$ ($s=2$ in the above figure), forming a gathered feature sequence. 
        [Step 3] \textit{Predicted TSM} is computed by calculating a pairwise similarity between the gathered feature sequence and the \textit{projected} gathered feature sequence. Position-wise BCE loss is then posed with the target TSM and the predicted TSM.
        }
\label{fig:model}
\end{figure}

\subsection{Overview of Soft-Landing Strategy}
\label{sec:sola_module}
The main idea of the Soft-Landing (SoLa) strategy is placing a light-weight neural network, or SoLa module, between the pretrained encoder and the downstream head.
In our framework, there is no need for retraining or fine-tuning of the pretrained encoder since the SoLa module is solely responsible for narrowing the task discrepancy gap between the pretrained encoder and the downstream head.
Besides, following the standard two-stage TAL convention, our SoLa strategy works on the snippet feature level and hence eliminates the need for an access to raw RGB frames.
As most properties of the standard TAL framework are preserved, our SoLa strategy requires only a \textit{minimal modification} of the existing framework for its integration.

For a clear demonstration, overall schematic diagram of the training and the inference stage of the SoLa strategy is illustrated in Figure~\ref{fig:model} and Figure~\ref{fig:teaser} (b) respectively.
In the training stage, we first sample a fixed-size local subsequence from the pre-extracted snippet features $f_{\tau =1}^{L}$, which originate from the pretrained snippet encoder.
Next, the sampled subsequence is put into the SoLa module, yielding a transformed feature sequence.
After assembling each element with a step size $s$, the shortened feature sequence is projected through an MLP and pairwise similarities between the sequence before the projection and after the projection are computed, forming a TSM-like matrix.
In Figure~\ref{fig:model}, we denote this asymmetric self similarity matrix as a predicted TSM.
With a predefined target TSM which solely depends on the frame interval, Similarity Matching loss $\mathcal{L}^{\mathrm{SM}}(\cdot, \cdot)$ is posed in each position of the matrices.
Note that the above training procedure does not demand any additional label as the frame interval (distance between the frames) is its only supervisory signal.
The exact procedure for generating the target TSM and calculating the loss will be discussed in the next section. (Section~\ref{sec:sm})

During the inference stage, we put the snippet feature sequence $f_{\tau =1}^{L}$ to the trained SoLa module $SoLa(\cdot)$ and get an enhanced feature sequence $F_{\tau =1}^{L}$.
Then, $F_{\tau =1}^{L}$ is directly used for the downstream task.

From its definition, the SoLa module can be any neural networks that take a tensor with shape $(L,m)$ as its input and output a tensor of the same shape.
To demonstrate it as a proof of concept, we employed the simplest architecture: a shallow 1D CNN.
While devising effective architecture for the SoLa module is an interesting future research direction, we found that this 1D CNN architecture works surprisingly well.
More detailed architectural configurations and ablation studies on this topic are provided in the supplementary materials.

\begin{figure}[t]
        \includegraphics[width=\linewidth]{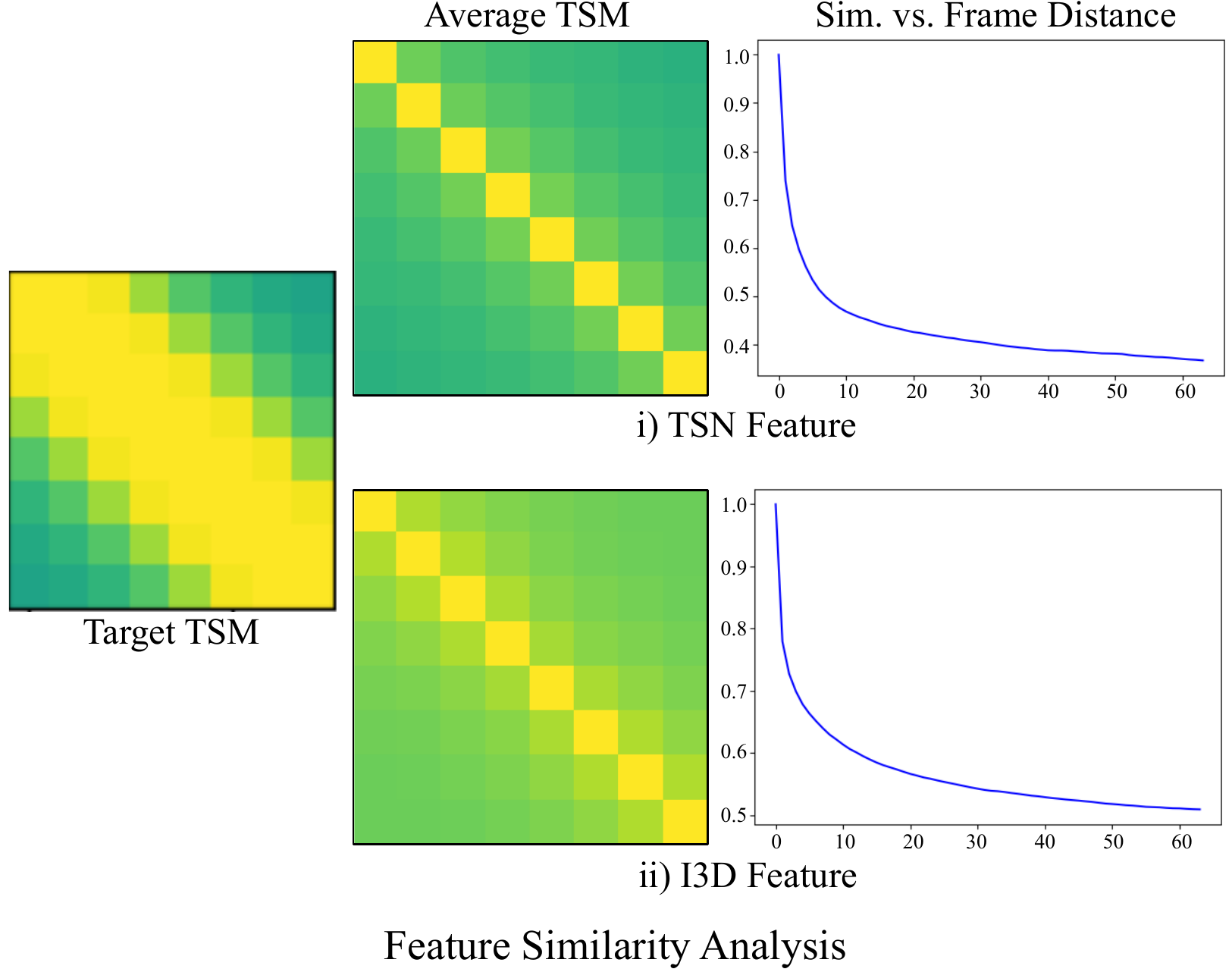}
    \caption{
         (Right) Results of the feature similarity analysis. Two different snippet encoders (\cite{wang2016temporal, carreira2017quo}) are utilized for extracting snippet features. For each video, snippet feature subsequence with a fixed length is sampled and a TSM is computed for each of those subsequences by calculating their pairwise self-similarities. The average over those TSMs is denoted as the ``Average TSM''.\\
         (Left) Target TSM with $K=16$. Target TSM exaggerates the temporal similarity structure of the average TSMs.
        }
\label{fig:feature_similarity_analysis}
\end{figure}

\subsection{Similarity Matching}
\label{sec:sm}
Due to the \textit{unlabeled} target dataset assumption, it is obvious that the training of the SoLa module must be done in a self-supervised way.
While recent studies~\cite{chen2021exploring, grill2020bootstrap} have shown remarkable success of contrastive learning in the self-supervised representation learning domain, all of these methods rely on extensive data augmentation.
Specifically, \cite{chen2020simple} pointed out that a strong data augmentation is essential for successful training of the self-supervised contrastive learning models.
Nevertheless, existing data augmentation methods are rgb frame-level operation (e.g. random cropping, color distortion, etc.), whereas our SoLa module deals with feature sequences $f_{\tau =1}^{L}$.
Since feature-level data augmentation is infeasible, straightforward application of previous contrastive learning approach is non-trivial.

Instead of designing a feature-level data augmentation technique, we pay attention to the \textit{temporal similarity structure} that general videos naturally convey: ``Adjacent frames are similar, while remote frames remain distinct.''
This intuition is validated in Figure~\ref{fig:feature_similarity_analysis}, where the similarities among the snippet features from the same video and their average Temporal Self-similarity Matrix (TSM) are plotted. 

As expected, the frame similarity decays as the frame interval increases, regardless of the backbone snippet encoder's architecture.
Note that although specific similarity scores may vary between the backbone architectures, the overall tendency of the similarity decay and the average TSM is preserved, indicating that the temporal similarity structure is a common characteristic of general videos.
With this empirical observation, we propose a novel pretext task for feature level self-supervised pretraining called  \textit{Similarity Matching}, which utilizes frame interval as its only learning signal.

One of the possible approaches for exploiting the frame distances to train the model is directly using their positional information ~\cite{yun2022time, zhai2022position}.
However, instead of using the raw frame interval as the learning signal, we go one step further to produce a snippet feature sequence that is even more \emph{temporally sensitive}.
To this end, we introduce a deterministic function $\lambda(\cdot): \mathbb{R} \to \mathbb{R}$ that maps a frame interval to a desirable, i.e., target similarity score.
With these scores, the SoLa module is trained with the Similarity Matching.
Here, the Similarity Matching denotes a direct optimization of the feature similarity; if the target similarity score is given as 0.75, the objective function would penalize the feature encoder if the similarity score between the two snippet features deviates from 0.75.
In this manner, the objective forces the similarity between the two snippet features to be close to the target similarity score, which depends on the frame interval between them.

We designed the $\lambda$ function to enhance the temporal similarity structure of the empirical similarity score distribution.
To achieve the goal, the function should make the adjacent snippet features be \textit{more similar} while remote snippet features remain distinct.
To do so, the $\lambda$ function is defined as follows with a constant $K$:
\begin{equation}
\label{eq:lambda}
    \lambda(d)=\sigma \Big(\frac{K}{d^2} \Big), d \neq 0,
\end{equation}
where $d$ stands for the \textit{frame interval} between the two snippet representations and $\sigma(\cdot)$ denotes the sigmoid function.
High $K$ value in Equation~\eqref{eq:lambda} leads to a $\lambda$ function that amplifies the similarities of neighboring snippet features (refer to the Target TSM and the Average TSM in Figure~\ref{fig:feature_similarity_analysis}).
We found that this target assignment is important for achieving a good downstream performance (see ablation study results in the supplementary materials).
Furthermore, it is worth mentioning that the target similarity score never goes to zero with the above $\lambda$ function if both snippet features are from the same video, reflecting our another prior knowledge: ``There is an invariant feature that is shared among frames in the same video'', which also corresponds to the basic concept of the slow feature analysis~\cite{wiskott2002slow}.

Remaining design choice for the Similarity Matching is choosing an appropriate way to produce a similarity prediction $\hat{p}$ from the given snippet feature pair $(f_i, f_j)$.
As the frame interval between different videos cannot be defined and to avoid trivial dissimilarity, we presume the snippet feature pair $(f_i, f_j)$ to come from the same video.

Motivated by the Simsiam framework~\cite{chen2021exploring}, we adopt asymmetric projector network $Proj(\cdot): \mathbb{R}^m\to\mathbb{R}^m$, which consists of two fully-connected layers.
Thus, we first calculate  $z_i=SoLa({f_{\tau =1}^{L}})[i], z_j=Proj(SoLa(f_{\tau =1}^{L})[j])$ and utilize the rescaled cosine similarity between them as the network prediction:
\begin{equation}
\label{eq:rescaled_cosine}
    \hat{p}= \frac{1}{2}(\frac{z_i \cdot z_j}{\parallel z_i \parallel \parallel z_j \parallel}+1).
\end{equation}

Finally, with those $\hat{p}$ and target similarity $\Lambda$ (output from the $\lambda$ function in Equation~\eqref{eq:lambda}), Similarity Matching loss $\mathcal{L}^{\mathrm{SM}}$ is computed as follows:
\begin{equation}
\label{similaritymatching}
    \mathcal{L}^{\mathrm{SM}}(\Lambda, \hat{p})=-\Lambda\log \hat{p}-(1-\Lambda)\log(1-\hat{p}).
\end{equation}
Note that $\mathcal{L}^{\mathrm{SM}}(\cdot, \cdot)$ is merely a standard Binary Cross Entropy (BCE) loss with a soft target.
In the supplementary materials, we provide the connection between the Similarity Matching loss and the contrastive learning.

Focusing on the fact that both $\Lambda$ and $\hat{p}$ values represent \textit{feature similarities}, we can derive an interesting interpretation: view the set of $\Lambda$ and $\hat{p}$ values as TSMs.
In this point of view, posing standard BCE loss between $\Lambda$ and $\hat{p}$ values becomes a TSM Matching.
The target TSM, presented in Figure~\ref{fig:feature_similarity_analysis}, visually illustrates $\Lambda$ assignments following Equation~\eqref{eq:lambda}; it can be seen as a general and desirable TSM of an arbitrary video -- maintaining high similarity between close frames while low similarities among distant frames.
For each $\hat{p}$ and $\Lambda$ in the corresponding position, Equation~\eqref{similaritymatching} is computed, resulting in a group of BCE losses.
This group of BCE losses can be succinctly described as an aforementioned ``TSM matching'' in that the losses force the predicted TSM to resemble the target TSM as training proceeds.

\begin{figure}[t]
        \includegraphics[width=\linewidth]{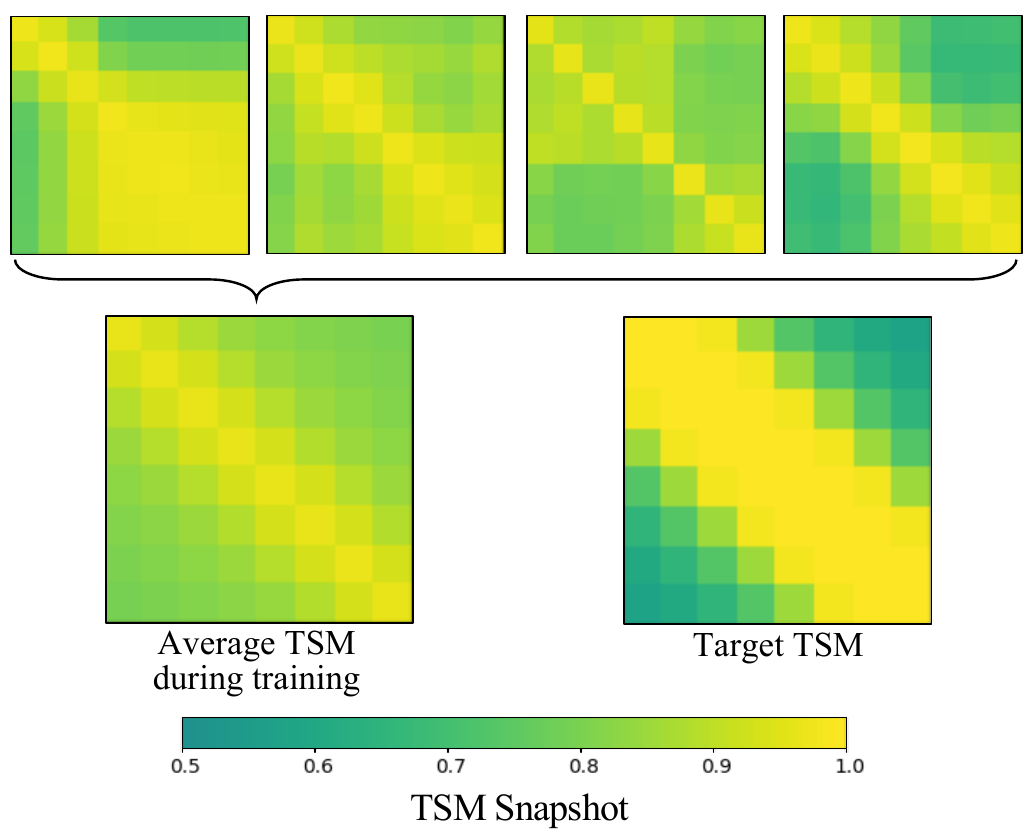}
    \caption{Snapshot of TSMs during training. Average TSM here is computed with samples in 1 batch (256 snippet feature subsequences)
    }
\label{fig:training_snapshot}
\end{figure}

One may suspect that the TSM matching with a fixed target TSM would induce monotonous TSM prediction.
However, we observed that only the \textit{average} of the predicted TSMs goes to the target TSM as training proceeds while each sample shows diverse patterns (Figure~\ref{fig:training_snapshot}).

\begin{table*}[t]
    \centering
    \resizebox{\linewidth}{!}{
        \renewcommand{\arraystretch}{1.3}
        \begin{tabular}{c|c|c|c|cccc>{\columncolor[gray]{.9}}c|cccc>{\columncolor[gray]{.9}}c|c|c}
            \hlineB{3}
            \multirow{2}{*}{\textbf{Method}}    & \multirow{2}{*}{\textbf{Backbone}}   & \multirow{2}{*}{\textbf{T.E}}   & \multirow{2}{*}{\textbf{Label}}  
                                                    & \multicolumn{5}{c|}{\textbf{Temporal Action Localization (GTAD\cite{xu2020gtad})}}    & \multicolumn{5}{c|}{\textbf{Temporal Action Proposal (BMN\cite{lin2019bmn})}}    & \multirow{2}{*}{\shortstack{\textbf{Required} \\ \textbf{Memory}}} & \multirow{2}{*}{\shortstack{\textbf{Flops} \\ \textbf{(per clip)}}}           \\ 
                                                            &&&& mAP@0.5    & @0.75    & @0.95         & Avg.           & Gain          & AR@1          & @10            & @100           & AUC            & Gain   &&\\ \hline
            \multirow{5}{*}{Baseline}
                             & TSM-R18\cite{lin2019tsm}
                                            & -     & -     & 49.64         & 34.16    & 7.68          & 33.59         & -        & 33.29         & 56.20     & 74.88     & 66.81     & -      & -        & - \\ 
                             & TSM-R50\cite{lin2019tsm}
                                            & -     & -     & 50.01         & 35.07    & 8.02          & 34.26         & -        & 33.45         & 56.55     & 75.17     & 67.26     & -      & -        & - \\ 
                             & I3D\cite{carreira2017quo}
                                            & -     & -     & 48.50         & 32.90    & 7.20          & 32.50         & -        & 32.30         & 54.60     & 73.50     & 65.60     & -      & -        & - \\ 
                             & R(2+1)D-34\cite{tran2018closer}
                                            & -     & -     & 49.76         & 34.87    & 8.65          & 34.08         & -        & 34.67         & 57.89     & 75.65     & 68.08     & -      & -        & - \\
                             & TSN\cite{wang2016temporal}
                                            & -     & -     & 49.78         & 34.46    & 7.96          & 33.84         & -        & 33.59         & 56.79     & 75.05     & 67.16     & -      & -        & - \\ \hline
            \multirow{2}{*}{BSP\cite{xu2021boundary}}
                             & TSM-R18      & \cy   & \cc   & 50.09         & 34.66    & 7.95          & 33.96         & +0.37    & -             & -         & -         & -         & -      & \multirow{2}{*}{N/A}      & 14.6G\\ 
           
                             & TSM-R50      & \cy   & \cc   & 50.94         & 35.61    & 7.98          & 34.75         & +0.49    & 33.69         & 57.35     & 75.50     & 67.61     & +0.35  &          & 33G \\ \hline
            \multirow{2}{*}{LoFi\cite{xu2021low}}
                             & TSM-R18      & \cy   & \cy   & 50.68         & 35.16    & 8.16          & 34.49         & +0.90    & 33.71         & 56.81     & 75.58     & 67.49     & +0.58  & \multirow{2}{*}{128G} & 14.6G \\ 
            
                             & TSM-R50      & \cy   & \cy   & 50.91         & 35.86    & 8.79          & 34.96         & +0.70    & -             & -         & -         & -         & -      &          & 33G \\ \hline
            PAL\cite{zhang2022unsupervised}
                             & I3D          & \cy   & \cx   & 49.30         & 34.00    & 7.90          & 33.40         & +0.90    & 33.70         & 55.90     & 75.00     & 66.80     & +1.24  & 2048G    & 3.6G \\ \hline
            TSP\cite{alwassel2021tsp}
                             & R(2+1)D-34   & \cy   & \cy   & 51.26         & 36.87    & 9.11          & 35.81         & +1.73    & 34.99         & 58.96     & 76.63     & 69.04     & +0.96  & 64G      & 76.4G \\ \hline
            \textbf{SoLa(ours)}       & TSN       & \cx   & \cx   & 51.17         & 35.70    & 8.31          & 34.99         & +1.15    & 34.25         & 57.75     & 75.86     & 68.07     & +0.91  & \textbf{11G}      & \textbf{0.014G}\\    \hlineB{3}
        \end{tabular}
    }
    \caption{TAL results on ActivityNet-v1.3 dataset. T.E stands for the ``Trainable snippet Encoder''. \cy, \cc, \cx \  in Label column denote ``Temporal annotation'', ``Action class annotation'', and ``No label at all'' respectively. ``Memory'' refers the GPU memory constraint, which is reported according to the hardware configuration of each method. \cite{xu2021boundary} does not provide its hardware configuration. Per clip FLOP values of other methods are from the main table of~\cite{zhang2022unsupervised}.}
    \label{tab:gtad_table}
\end{table*}

\section{Experiments}
\label{sec:experiments}
\subsection{Experimental Setup}
\paragraph{Target Downstream Tasks}
To validate the representation power of the transformed feature sequences, we evaluated TAL performance with off-the-shelf downstream heads including G-TAD~\cite{xu2020gtad}, BMN~\cite{lin2019bmn}, and Actionformer~\cite{zhang2022actionformer}.
Moreover, we tested our transformed feature sequence in the language grounding task with the LGI~\cite{mun2020local} head.
Result from Actionformer~\cite{zhang2022actionformer} downstream head is presented in our supplementary materials.
For an additional experiment, we also present a linear evaluation result, which aims to distinguish foreground snippet features from background snippet features with an one-layer classifier.
Following the related works~\cite{wu2018unsupervised, xu2021boundary, xu2021low,alwassel2021tsp}, we choose G-TAD~\cite{xu2020gtad} as the main downstream head and conducted various ablation studies.
For all training procedures of the downstream tasks, we followed the same procedure of the published codes.
No additional fine-tuning of the hyperparameters in the downstream heads is done.

\paragraph{Dataset and Feature}
ActivityNet-v1.3~\cite{caba2015activitynet} dataset consists of 19,994 temporally annotated, untrimmed videos with 200 action classes.
Videos in the dataset are divided into training, validation, and test sets by the ratio of 2:1:1. 
For ActivityNet-v1.3 SoLa training, we used the training split with \textit{no label}.

HACS1.1 dataset~\cite{zhao2019hacs} is a newly introduced video dataset for the temporal action localization task. It  contains 140K complete action segments from 50K untrimmed videos with 200 action categories, which correspond to the ActivityNet-v1.3 action categories.
THUMOS14~\cite{idrees2017thumos} contains 413 untrimmed videos with 20 action classes and it is split into 200 training videos and 213 test videos.
Analogous to ActivityNet-v1.3 setting, we used the training splits of the aforementioned datasets with \textit{no label} for training the SoLa module.

Charades-STA~\cite{gao2017tall} is a popular video grounding dataset which is an extension of the action recognition dataset called Charades~\cite{sigurdsson2016hollywood}.
It contains 9848 videos of daily indoors
activities, with 12408/3720 moment-sentence pairs in train/test set respectively.

For ActivityNet-v1.3 and THUMOS14 experiments, we chose Kinetics400~\cite{kay2017kinetics} pretrained two-stream TSN network~\cite{wang2016temporal} as the frozen snippet encoder, which is widely used among various TAL methods.
In this setting, the sampling step size of the snippets is set to 5.

For HACS1.1 experiments, we directly utilized the officially available\footnote{http://hacs.csail.mit.edu/challenge.html} I3D~\cite{carreira2017quo} feature, which is extracted from Kinetics400~\cite{kay2017kinetics} pretrained I3D snippet encoder.
For all the SoLa training procedure, we used \textit{pre-extracted} snippet feature sequences (numpy array format), indicating that there is no parameter update on the snippet encoders.

For the VG task, we extracted I3D features of Charades-STA~\cite{gao2017tall} dataset, using Kinetics400~\cite{kay2017kinetics} pretrained weight from the PytorchVideo model zoo.\footnote{https://github.com/facebookresearch/pytorchvideo/}

\paragraph{SoLa Settings}
Except for the SoLa configuration for the unified training in the ablation study, the training and the evaluation of SoLa follows a two-staged approach.
First, the SoLa module is trained with the Similarity Matching and with the trained module, input feature sequences $f_{\tau =1}^{L}$ are converted to the transformed feature sequences $F_{\tau =1}^{L}$.
Then, the training and the evaluation of the downstream task are conducted on the transformed feature sequence $F_{\tau =1}^{L}$.
For all TAL tasks, we set the output dimension of the SoLa module to be the same as its input dimension, yielding identical tensor shape for the input feature sequences $f_{\tau =1}^{L}$ and the transformed feature sequences $F_{\tau =1}^{L}$.

Unlike previous works that required substantial amount of computational resources for training snippet encoders, we only used 1 RTX-2080 GPU for training SoLa.
In fact, only less than 6GB of GPU memory and approximately 3 hours for training are required for the SoLa training.
Other hyperparameter settings can be found in the supplementary materials.

\paragraph{Evaluation Metrics}
Following the standard evaluation protocols, we report mean Average Precision (mAP) values under different temporal Intersection over Union (tIOU) thresholds for G-TAD performance. An average mAP is also reported by averaging across several tIoUs for clear demonstration.
For BMN, we adopted the standard AR@k (the average recall of the top-k predictions) and AUC (the area under the recall curve) as performance metrics.
For VG tasks, the top-1 recall score at three different tIoU thresholds $\{0.3, 0.5, 0.7\}$ are presented.
Following LGI’s evaluation metrics~\cite{mun2020local}, mean tIoU between predictions and ground-truth is also reported.

\begin{table}
    \centering
    \resizebox{\linewidth}{!}{
    \renewcommand{\arraystretch}{1.3}
    \begin{tabular}{c|c|ccccc}
        \hlineB{3}
        \multirow{2}{*}{\textbf{Datasets}}    & \multirow{2}{*}{\textbf{Method}}
        & \multicolumn{5}{c}{\textbf{Temporal Action Localization (GTAD\cite{xu2020gtad})}} \\
                        &               & mAP@0.5   & @0.75     & @0.95     & \multicolumn{2}{c}{Avg}   \\ \hline
        \multirow{2}{*}{HACS\cite{zhao2019hacs}}
                        & Baseline      & 40.45     & 26.84     & 8.10      & \multicolumn{2}{|c}{26.85} \\ \cline{2-7}
                        & \textbf{SoLa(ours)}    & 41.12     & 27.69     & 8.54      & \multicolumn{2}{|c}{27.58} \\ \hline\hline
        \multicolumn{2}{c|}{ }          & mAP@0.3   & @0.4   & @0.5   & @0.6    & @0.7    \\ \hline
        \multirow{2}{*}{THUMOS\cite{idrees2017thumos}}
                        & Baseline      & 57.93     & 51.96     & 43.05     & 32.62     & 22.52 \\ \cline{2-7}
                        & \textbf{SoLa(ours)}    & 59.14     & 53.08     & 44.32     & 33.59     & 23.19 \\ \hline
    \end{tabular}}
    \caption{TAL results on HACS~\cite{zhao2019hacs} and THUMOS14~\cite{idrees2017thumos} datasets. In both datasets, SoLa module brings downstream performance gain, supporting the general applicability of the SoLa strategy.}
    \label{tab:other_datasets}
\end{table}

\begin{table}
    \centering
    \resizebox{\linewidth}{!}{
    \renewcommand{\arraystretch}{1.3}
    \begin{tabular}{c|ccccc>{\columncolor[gray]{.9}}c}
        \hlineB{3}
        \multirow{2}{*}{\textbf{Method}}
        & \multicolumn{6}{c}{\textbf{Video Grounding (LGI)}} \\
                        & R@0.1     & R@0.3     & R@0.5     & R@0.7     & mIoU  & gain \\ \hline\hline
        Baseline        & 67.72     & 56.09     & 41.48     & 21.88     & 38.24 & -       \\ \hline
        SoLa            & 70.50     & 58.47     & 42.96     & 21.90     & 39.70 & +1.46 \\ \hlineB{3}
    \end{tabular}}
    \caption{VG task performance (LGI). mIoU gain is reported.}
    \label{tab:lgi}
\end{table}

\subsection{Main Results}
As a relatively new task, many of the preceding works did not release their codes yet, making the exact fair comparison difficult.
Thus, we will mainly focus on the \textbf{gain}, which reflects given method's efficacy compared to their baselines.

\paragraph{TAL Results}
Table\ref{tab:gtad_table} and Table~\ref{tab:other_datasets} show that our SoLa module not only achieves exceptional performance compared to most of the recent snippet encoder training approaches, but also generalizes well to various datasets and different snippet encoders.
Note that TSP~\cite{alwassel2021tsp} heavily relies on temporally annotated video dataset to train the snippet encoder.
In addition, our framework has the general applicability across various downstream tasks, as evidenced by the performance gain on VG task in Table~\ref{tab:lgi}.
These results allude that when it comes to bridging the task discrepancy gap, computationally heavy snippet encoder pretraining is not an essential part.
This counterintuitive results strongly suggest that developing SoLa strategy can be a promising research direction.

\begin{figure}[t]
        \includegraphics[width=\linewidth]{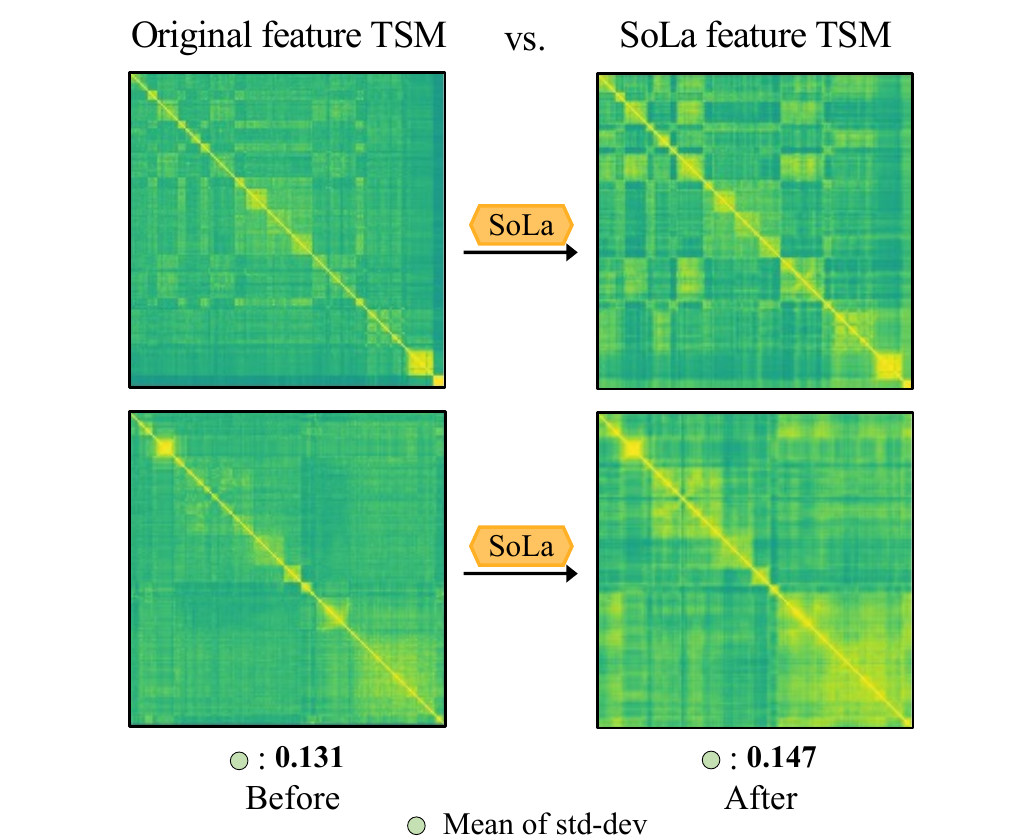}
    \caption{
        Qualitative results showing the effect of the SoLa module which compensates the temporal sensitivity. 
        For each video in Activitynet1.3~\cite{caba2015activitynet} validation split, standard deviation of the pairwise self-similarity scores is calculated, and the average value of them is reported as the ``Mean of std-dev''.
    }
\label{fig:qualitative_result}
\end{figure}

\paragraph{Computational Cost}
As our method directly processes the snippet feature and does not involve snippet encoder training, it is exceptionally efficient in computation.
It is quantitatively verified with the memory requirement and per clip FLOP in Table~\ref{tab:gtad_table}.
Here, per clip FLOP values of the other methods, which are from the main table of~\cite{zhang2022unsupervised}, only represent the backbone network's computational complexity; they are lower bounds of computational cost with regard to their whole training procedure since they involve backbone network training.

\paragraph{Qualitative Results}
Figure~\ref{fig:qualitative_result} visually illustrates the role of the trained SoLa module.
We can see that TSMs of the SoLa feature sequences show more diverse patterns compared to the original ones.
To support the claim, we calculated the average value of standard deviation.
Standard deviation here is computed for each snippet feature sequence's self similarity scores, meaning that higher standard deviation is an indication of more temporally sensitive feature sequence.
We observed that SoLa features sequences are more temporally sensitive.

\paragraph{Linear Evaluation}
The most straightforward and standard method of measuring the learned feature's representation power is conducting a linear classification on them~\cite{chen2020simple, grill2020bootstrap}.
It is based on the linear separability concept - \textit{good features should be linearly separable for the meaningful criterion.}
As we are targeting on the TAL tasks and our Similarity Matching training compensates the temporal sensitivity of the snippet features in the same video, we chose the foreground/background frame distinction as a criterion for the linear evaluation protocol, forming a binary classification problem.
Both the original and the transformed features are normalized before they are fed into the linear classifier.

\begin{figure}[t]
    \includegraphics[width=0.8\linewidth]{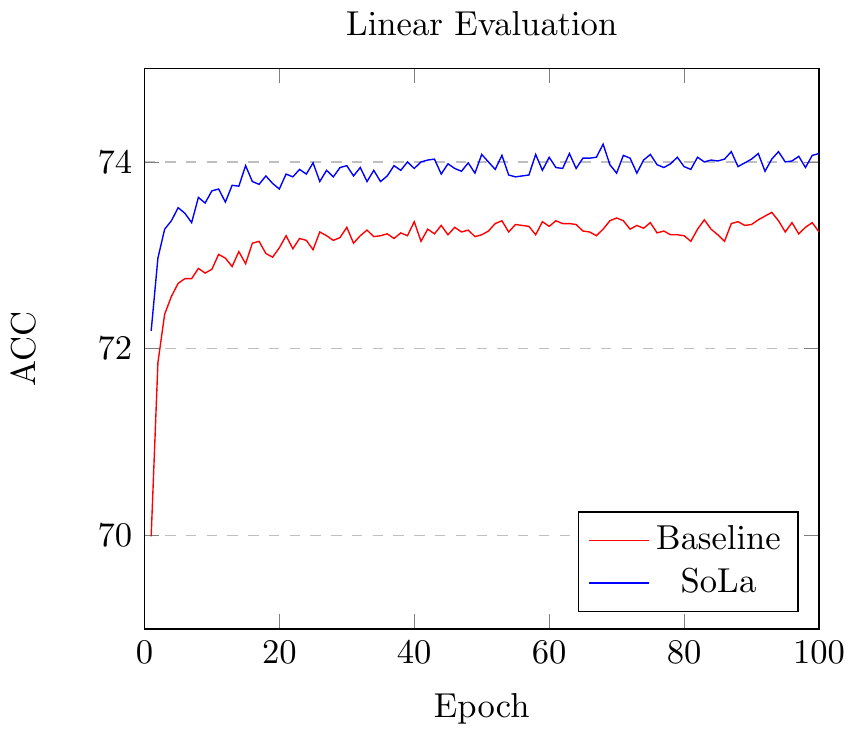}
    \caption{Linear evaluation performance graph in Activitynet 1.3. Test accuracy is reported at the end of each epoch.}
\label{fig:linear_eval}
\end{figure}

We present linear evaluation results in  Figure~\ref{fig:linear_eval}.
It clearly shows that the SoLa module with Similarity Matching improves the temporal sensitivity, making the linear classifier easier to distinguish foreground/background snippet.

\subsection{Ablation Study}

We conducted three major ablation studies on the SoLa module with the G-TAD downstream head by experimenting with the followings: i) receptive fields, ii) unified training of the SoLa module, iii) different snippet encoder which only uses RGB-only frames.

As we discussed in Section~\ref{sec:sola_module}, the SoLa module consists of simple 1D convolutional layers.
However, finding the best-performing convolution kernel size  remains as a design choice.
To figure out the influence of the design choice, we tested \verb +kernel_size+  $\in \{1,3,5,7\}$.

Moreover, a recent work on MLP projector~\cite{wang2021revisiting} claims that additional MLP between the classifier head and the feature encoder can reduce the transferability gap, inducing better downstream task performances.
From this, it is natural to ask if the SoLa module itself (without Similarity Matching) can play a role as a \textit{buffer} between the pretrained encoder and the TAL heads.
To answer this question, we tried a unified training of the SoLa module where the SoLa module is directly attached to the very front part of the TAL head and trained simultaneously with the head.
It can be seen as a TAL head with an additional local aggregation module on its front part, or a truncated version of~\cite{xu2021low}.

The prevailing convention in TAL is using the twostream network.
However, to ensure our SoLa strategy's general applicability, we also conducted experiments on RGB only I3D~\cite{carreira2017quo} features\footnote{The feature is available in https://github.com/Alvin-Zeng/PGCN} 

Table~\ref{tab:anet_ablation} describes the result of ablation studies on our SoLa strategy.
The result suggests that while the local feature aggregation plays a key role in the SoLa module, the wider receptive field does not necessarily entail a performance improvement; we attribute it to the need for an appropriate information bottleneck since the frame interval is exploited as the model's learning signal.
In addition, we find out that while the SoLa module itself can slightly boost the TAL performance, it does not match the performance of the SoLa module trained with Similarity Matching.
It implies that training with Similarity Matching is essential for the successful training of the SoLa module.

Lastly, we observed that the SoLa module performs reasonably well in the RGB-only case (bottom part in Table~\ref{tab:anet_ablation}).
However, the gain is lower than the two-stream case, indicating that there is an inherent limitation that comes from the frozen snippet encoder assumption.
For instance, if the base feature does not contain sufficient information, it serves as an information bottleneck for the SoLa module, imposing an upper bounds of the gain that SoLa module can achieve.
For more ablation studies and results about our model, please refer to our supplementary materials.


\begin{table}
    \centering
    \resizebox{\linewidth}{!}{
    \renewcommand{\arraystretch}{1.3}
    \begin{tabular}{c|cccc}
        \hlineB{3}
        \textbf{Setting}& \multicolumn{4}{c}{\textbf{Temporal Action Localization (GTAD\cite{xu2020gtad})}} \\
        Kernel         & mAP@0.5   & @0.75     & @0.95     & Avg   \\ \hline\hline
        Baseline        & 49.78     & 34.46     & 7.96      & 33.84 \\ \hline
        1               & 50.30     & 34.90     & 8.52      & 34.30 \\
        3               & 50.70     & 35.39     & 8.14      & 34.68 \\
        \hspace{4.7ex}5\textsubscript{Unified}
                        & 50.07     & 34.86     & 7.07      & 34.07 \\
        \rowcolor{gray!=25}
        5               & 51.17     & 35.70     & 8.31      & 34.99 \\
        7               & 50.94     & 35.53     & 7.53      & 34.79 \\ \hline\hline
        Baseline\textsubscript{I3D}
                        & 49.11     & 33.70     & 7.49      & 33.18 \\ \hline
        \rowcolor{gray!=25}
        \hspace{2.5ex}5\textsubscript{I3D}               & 49.86     & 34.41     & 6.29      & 33.59 \\ \hlineB{3}
    \end{tabular}}
    \caption{Ablation study results on ActivityNet-v1.3. The subscript \textit{Unified} refers to the simultaneous training of the SoLa module with the standard TAL head, where the module is trained with the standard downstream objective, not with our Similarity Matching. }
    \label{tab:anet_ablation}
\end{table}
\section{Conclusion}
In this paper, we introduced the Soft-Landing (SoLa) strategy, a new direction in tackling the suboptimal snippet encoder problem in TAL research.
Unlike previous works, we adopt a light-weight Soft-Landing (SoLa) module between the frozen encoder and the TAL head, resulting in both efficiency and easier applicability.
Coupled with the novel self-supervising method called the Similarity Matching, our SoLa strategy brings about significant performance gains in downstream tasks, outperforming most of the recent works that involve retraining the snippet encoder.
We hope that our results spark further research on efficient and effective methods of using the pretrained snippet encoder for long-form video understanding.

{\small
\bibliographystyle{ieee_fullname}
\bibliography{egbib}
}

\end{document}